%% file: main.tex
\documentclass{article}

\usepackage[main, preprint]{neurips_2026}
\usepackage[utf8]{inputenc} % allow utf-8 input
\usepackage[T1]{fontenc}    % use 8-bit T1 fonts
\usepackage{hyperref}       % hyperlinks
\usepackage{url}            % simple URL typesetting
\usepackage{booktabs}       % professional-quality tables
\usepackage{amsfonts}       % blackboard math symbols
\usepackage{nicefrac}       % compact symbols for 1/2, etc.
\usepackage{microtype}      % microtypography
\usepackage{xcolor}         % colors
\usepackage{multirow}
\usepackage{graphicx}
\usepackage{colortbl}
\usepackage{amsmath}

\usepackage{subcaption}
\usepackage{xcolor}
\usepackage{booktabs}

\input{defs}

\title{The Unreasonable Effectiveness of VLMs for Zero-shot Procedural Mistake Detection}

\author{
Serdar Ozsoy\textsuperscript{*1,3} 
\And
Lars Doorenbos\textsuperscript{*1,3} 
\And
Federico Spurio\textsuperscript{1,3} 
\And
Gianpiero Francesca\textsuperscript{2} 
\And
Juergen Gall\textsuperscript{1,3} \\
\\
\textsuperscript{1}University of Bonn \quad
\textsuperscript{2}Toyota Motor Europe \\
\textsuperscript{3}Lamarr Institute for Machine Learning and Artificial Intelligence \\
\\
\texttt{\{soezsoy,doorenbos,spurio,gall\}@iai.uni-bonn.de} \\
\footnotesize\textsuperscript{*}Equal Contribution  \\
}

\begin{document}

\maketitle

\begin{abstract}

Procedural mistake detection is important for quality control and user assistance across many disciplines. Recent work in this field has achieved significant gains by using the reasoning capabilities of Video-Language Models (VLMs) as components within multi-stage pipelines, which consist of separate modules for supervised temporal action segmentation, error detection, and explainability. Consequently, they remain dependent on tailored training datasets and require task-specific training, limiting their wider applicability. To remedy this, we introduce zero-shot procedural mistake detection and propose a unified Zero-shot Procedural Mistake detection (\model) framework that jointly solves procedural mistake detection and temporal action segmentation with a single pre-trained VLM. 
By evaluating our framework on two canonical mistake detection benchmarks, EgoPER and CaptainCook4D, we find that \model~can perform these tasks successfully, while approaching, or even outperforming, the performance of fully supervised methods. For instance, we achieve a 4.4 point improvement in EDA and a 2.0 point improvement in F1@.5 on average over all five EgoPER tasks compared to the strongest supervised methods. 
Overall, our results show the potential of unified methods for procedural mistake detection, and we hope this will steer the field away from highly complex pipelines and toward more generally applicable solutions.

\end{abstract}

\input{sec/1_intro}

\input{sec/2_rw}
\input{sec/3_method}
\input{sec/4_exps}
\input{sec/6_conclusion}

% While promising, existing multi-stage methods lack cross-task transferability, produce non-interpretable outputs, and require threshold tuning. Therefore, we eliminate these limitations without requiring additional training by leveraging the extended context and natural visual reasoning capabilities of modern VLMs. 
% Procedural error detection in egocentric task videos is fundamentally a semantic reasoning problem, and sufficiently large pre-trained VLMs can solve this problem without domain-specific training. 

% \begin{ack}
% Use unnumbered first level headings for the acknowledgments. All acknowledgments
% go at the end of the paper before the list of references. Moreover, you are required to declare
% funding (financial activities supporting the submitted work) and competing interests (related financial activities outside the submitted work).
% More information about this disclosure can be found at: \url{https://neurips.cc/Conferences/2026/PaperInformation/FundingDisclosure}.

% Do {\bf not} include this section in the anonymized submission, only in the final paper. You can use the \texttt{ack} environment provided in the style file to automatically hide this section in the anonymized submission.
% \end{ack}

% \section*{References}

{
    \small
    \bibliographystyle{ieeenat_fullname}
    \bibliography{main}
}

% \newpage
%%%%%%%%%%%%%%%%%%%%%%%%%%%%%%%%%%%%%%%%%%%%%%%%%%%%%%%%%%%%

\appendix

\input{sec/7_appendix}

% \section{Technical appendices and supplementary material}
% Technical appendices with additional results, figures, graphs, and proofs may be submitted with the paper submission before the full submission deadline (see above). You can upload a ZIP file for videos or code, but do not upload a separate PDF file for the appendix. There is no page limit for the technical appendices. 

% Note: Think of the appendix as ``optional reading'' for reviewers. The paper must be able to stand alone without the appendix; for example, adding critical experiments that support the main claims to an appendix is inappropriate. 

%%%%%%%%%%%%%%%%%%%%%%%%%%%%%%%%%%%%%%%%%%%%%%%%%%%%%%%%%%%%

%\newpage
%\input{checklist.tex}

\end{document}

%% file: defs.tex
\definecolor{orange}{rgb}{1,0.5,0}
\definecolor{gr}{rgb}{0,0.65,0}
\definecolor{mygray}{gray}{0.95}
\definecolor{lgray}{gray}{0.9}

\newcommand{\model}{ZeProM}

\definecolor{actcolor}{RGB}{58, 126, 191}   % blue
\definecolor{actiocolor}{RGB}{200, 118, 43} % orange

\newcommand{\framerow}[5]{%
  \includegraphics[width=0.188\linewidth]{#1}\hfill
  \includegraphics[width=0.188\linewidth]{#2}\hfill
  \includegraphics[width=0.188\linewidth]{#3}\hfill
  \includegraphics[width=0.188\linewidth]{#4}\hfill
  \includegraphics[width=0.188\linewidth]{#5}%
}

\usepackage[most]{tcolorbox}

\newtcolorbox{promptbox}[1]{
    colback=gray!5, 
    colframe=gray!50, 
    arc=1mm, 
    boxrule=0.5pt, 
    left=2mm, right=2mm, top=2mm, bottom=2mm,
    fonttitle=\bfseries,
    coltitle=black,
    title=#1
}

%% file: sec/1_intro.tex
\section{Introduction}

Human errors are unavoidable in procedural tasks, yet lead to major losses in efficiency, productivity, and resources across a wide variety of domains (e.g.,~\cite{suliburk2019analysis}). As such, recognizing when and where a mistake has occurred is of crucial importance. In this area, dedicated deep learning methods have shown great potential. However, these methods are becoming increasingly complex, consisting of multiple modules in supervised pipelines to segment the input video and predict which segments are executed incorrectly, thereby limiting their broader applicability.

Unsurprisingly, recent works now turn to pre-trained Video-Language Models (VLM) in their procedural mistake detection methods due to their capability to reason about video segments without task-specific training.
So far, however, VLMs are only incorporated as small components within larger pipelines.
For instance, VLMs are used to generate scene graphs~\citep{guo2025procedural}, explain why segments are labeled as errors~\citep{flaborea2024prego}, or extract features of specific segments~\citep{lee2025error}.
Still, the zero-shot success of modern VLMs on novel tasks motivates exploring their efficacy for the entire procedural mistake detection task, especially as removing the reliance of existing methods on densely labeled videos for training, where even small modifications to the procedure require re-acquiring and re-labeling a training dataset, would greatly expand the domains in which mistakes can be detected. 
Therefore, we depart from the classic fully-supervised regime to introduce zero-shot procedural mistake detection and address the question: \textit{can existing VLMs perform procedural mistake detection in a zero-shot manner?}

\begin{figure}
    \centering
    \includegraphics[width=0.95\linewidth]{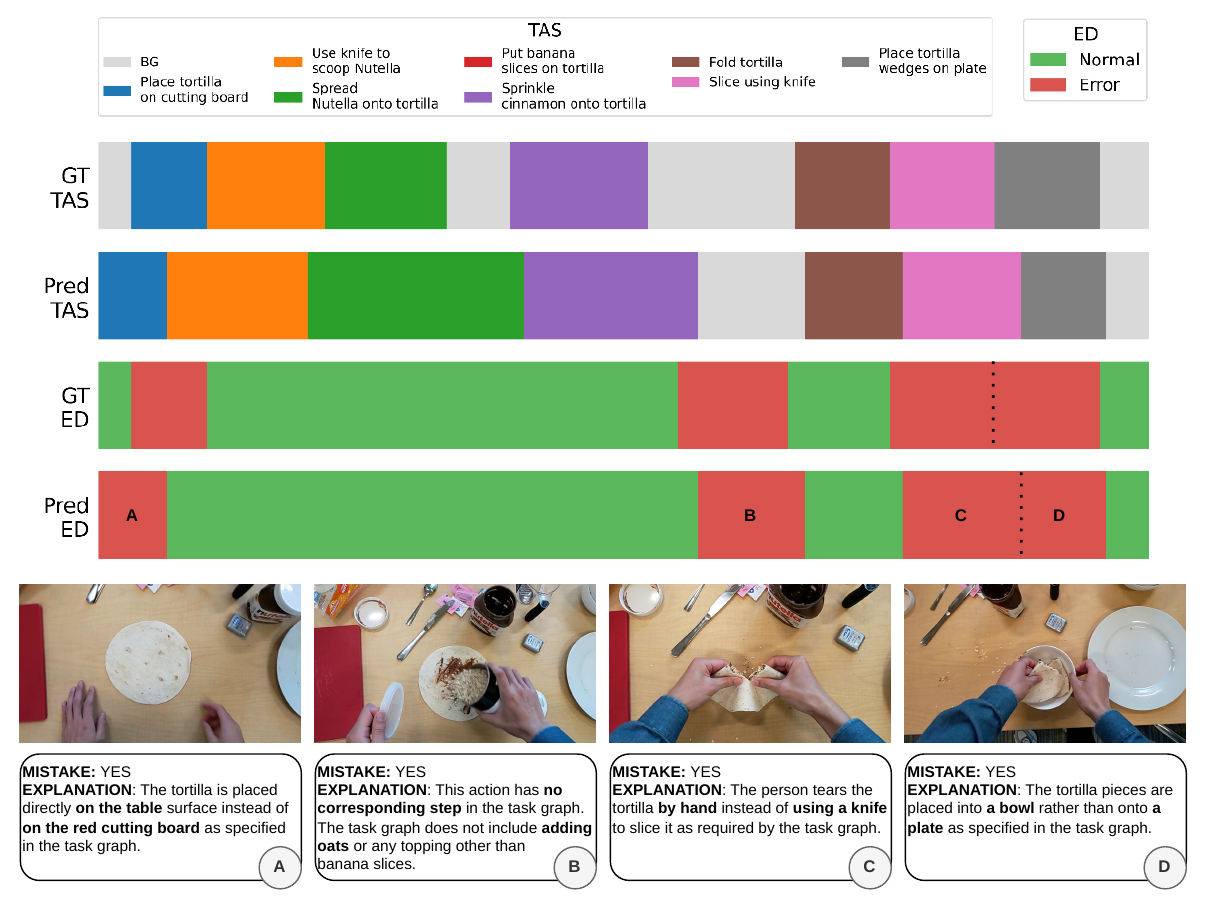}
    \caption{\textbf{Zero-shot procedural mistake detection with \model.} We show a video of the ``quesadilla" task from EgoPER. Without seeing any training samples, \model~successfully performs 1) temporal action segmentation, 2) mistake detection, and 3) error explanation, all within a single module.}
    \label{fig:teaser}
\end{figure}

To do so, we introduce our framework for Zero-shot Procedural Mistake detection (\model) with VLMs, and are the first to show that VLMs are capable of performing joint temporal action segmentation and mistake detection from only the task description, while obtaining mistake explanations as a byproduct for free.
\model~operates by segmenting the actions of the procedure and determining whether these segments are executed correctly, and performed in their correct order, as shown in Fig.~\ref{fig:teaser}, where it detects incorrect modifications to individual steps and an incorrectly added step with a single inference call.

To quantify its effectiveness, we evaluate \model~on two canonical procedural mistake detection tasks: EgoPER~\citep{lee2024error} and CaptainCook4D~\citep{peddi2024captaincook4d}, where we show that \model~performs surprisingly well. For example, \model~outperforms the state-of-the-art fully supervised methods by $4.4$ points in EDA and $2.0$ points in F1@.5, respectively, on segment-level mistakes of EgoPER. For EgoPER omission errors and CaptainCook4D mistakes, it approaches fully supervised methods up to within a few points.
In this way, our work advances the field by demonstrating the efficacy of zero-shot procedural mistake detection, offering strong training-free performance across tasks and matching or even outperforming fully supervised baselines. In contrast, previous methods are suitable for only one task and require training videos to learn their specifics, bringing additional computation and storage costs every time.
Overall, the main contributions of our work are:
\begin{itemize}
    \item We introduce zero-shot procedural mistake detection, in which models need to detect errors in procedural videos from the task instructions alone. 
    \item We propose \model, a simple and effective framework for joint zero-shot temporal action segmentation and procedural mistake detection, while obtaining error explanations for free.
    \item We show through experiments on established benchmarks that our framework comes close to, or even improves upon, fully supervised methods on standard benchmarks, despite not requiring any training data.
\end{itemize}

%% file: sec/2_rw.tex
\section{Related work}

Methods for procedural mistake detection can be roughly split into two categories: online and offline. Online methods process videos in a streaming manner and attempt to detect mistakes as they happen~\citep{ding2023every, flaborea2024prego, jang2019epic}.  
In this work, we focus on the more common category of offline methods. These use a video of the entire procedure to identify all segments corresponding to mistakes~\citep{lee2024error, ghoddoosian2023weakly, mazzamuto2025gazing, peddi2024captaincook4d, wang2023holoassist}.

Previous offline methods split the procedural mistake detection tasks into two main components: temporal action segmentation (TAS) and mistake detection. These two components are then tackled with specialized modules within a larger pipeline. Most commonly, this takes the form of a standard fully supervised TAS network followed by a module that labels detected segments as correct or erroneous~\citep{lee2025error,huang2025modeling}.
For instance, AEM~\citep{guo2025procedural} uses ActionFormer~\citep{zhang2022actionformer} to obtain segment features, then perform action effect modeling to determine which segments correspond to mistakes. 

Within these pipelines, VLMs have gained popularity to handle small parts of the task. For example, \cite{lee2025error} use them to generate possible mistake descriptions based on the procedural instructions, and \cite{guo2025procedural} use them to generate task graphs. 
\cite{flaborea2024prego} rely on VLMs to explain why a segment already labeled as a mistake would be one. 
As such, VLMs are typically only used in the mistake detection part of the pipeline, and these methods still rely on standard fully supervised networks for the temporal action segmentation. 
The single exception to this trend is the work of \cite{peddi2024captaincook4d}, who use VLMs for zero-shot error recognition. Here, the model has to categorize whether a single step contains an error, without focusing on its location in time. As a result, this setting is framed as a Video-Question-Answering task, for which the authors create segment-specific prompts with a large language model (LLM) to ask a VLM whether it was executed correctly. 
Our work differs in that, instead of asking whether a specific and predefined step is performed correctly, we identify every action step and determine which ones contain mistakes, jointly solving zero-shot temporal action segmentation and procedural mistake detection with a single module.

While zero-shot procedural mistake detection remains unexplored, VLMs have enabled zero-shot solutions to problems in related domains. For instance, they have shown great results in video anomaly detection~\citep{zanella2024harnessing, lee2025flashback}, out-of-distribution detection~\citep{ming2022delving,miyai2024generalized}, long video understanding~\citep{doorenbos2025video}, and so on. These successes form the motivation for \model, through which we show for the first time that their zero-shot capabilities translate to both TAS and mistake detection as well.

%% file: sec/3_method.tex
\section{Method}
\label{sec:method}

The goal of procedural mistake detection is to find segments within a video $V$ where the procedure execution deviates from instructions. 
This requires predicting a binary label for every frame $f$ in the video $V = \{f_i\}_{i=1}^{T}$, that denotes whether that frame corresponds to an error.
% This requires densely labeling every frame $f$ in the video $V = \{f_i\}_{i=1}^{T}$ to one of the $S$ instructional steps, a background label, or an error.
Previous works do this by training a temporal action segmentation (TAS) module to predict a label $\hat{y}_s \in \{a_0, a_1, \cdots, a_S\}$, with $a_0$ the background and $a_1$ to $a_S$ the procedural steps, for all frames $f$ in $V$. Then, a subsequent module is responsible for assigning segments to a binary error label $\hat{y}_e \in \{0, 1\}$, where $1$ denotes an error by convention. 
For instance, GTG2Vid~\citep{lee2025error} uses DiffAct~\citep{liu2023diffusion} to segment the video, then uses LLMs and VLMs to extract features and detect mistakes.
However, this approach comes with a reliance on densely labeled videos for training the TAS module. 
To alleviate this, we introduce zero-shot procedural mistake detection, in which models should predict 
error labels $\hat{y}_e$ for any given video from only its task description, i.e., the action descriptions and their dependencies.
We address this task by introducing \model, and describe its design below.

\subsection{\model: zero-shot procedural mistake detection}

\model~is a framework for zero-shot procedural mistake detection that does not require any training. Instead, the key choices that make our approach successful are the design of both the input and output representations. 

\noindent\textbf{Input representation.}
The main design choices for the input representation are the format of 1) the procedure description, and 2) the description of the task the model should perform.
For the former, previous works for procedural mistake detection rely on standard TAS methods that encode instructional steps as one-hot vectors. In contrast, \model~exploits the semantic information encapsulated in their descriptive names to perform a zero-shot temporal action segmentation. As the procedure description is now the only source of information our model can rely on, the way in which this information is provided to the model becomes important. This should clearly describe both the descriptions of individual actions, as well as the way in which they relate to each other. 

We opt for a simple list of numbered actions with their corresponding names to describe the procedure. We found that more complex ways of describing the procedure, for example, with their full graph structure, did not result in better performance. In particular, listing explicit dependencies between actions tends to confuse the model unnecessarily, as an early wrong prediction for a segment affects the entire structure negatively, which offsets the benefits of the full graph. 
% Therefore, we only pass the ordered procedure to the model. 
We justify this choice empirically in our ablation studies.

Given the information about the procedure, we need to instruct the VLM to perform the correct task.
Instead of asking the model to output the error labels $\hat{y_e}$ directly, we follow the combination of TAS and mistake detection common in previous fully supervised pipelines. Thus, we ask the model to perform a segmentation of the video into its actions, then detect which ones constitute mistakes. In this way, the model is encouraged to spend more computation (i.e., tokens) to further reason about the task, thereby increasing performance similar to chain-of-thought or reasoning methods~\citep{wei2022chain}. We provide our full prompt in the appendix.

\noindent\textbf{Output representation.}
A final important aspect is the structure of the output.
Rather than prompting the model to generate dense labels for every frame in the video, we instruct the model to generate a structured output in \texttt{JSON} format.
Specifically, we task the VLM to generate a list of segments, where every segment is a data structure with the segment ID, its start and end time, the action description, and the step $a_s$ to which it matches according to the model. Moreover, it contains a binary error field and a field to describe the error, which should only be populated if an error is detected. This way, \model~outputs an explanation of why certain segments are mistakes, leading to explainable results. We show examples of this in the next section.
We do not specify the number of segments in advance. Furthermore, we found that the models are powerful enough to generate non-overlapping, contiguous segments, and, therefore, do not apply any post-processing to the output.

\noindent\textbf{\model-MC: probabilistic mistake detection.}
By default, \model~assigns binary error labels to segments without probabilities. In some scenarios, however, probabilities can be helpful as an approximation of the confidence of the model. We found that directly prompting \model~to include a confidence score in its output did not provide satisfactory results. Instead, we additionally introduce \model-MC, which uses the randomness in the VLM token generation process to obtain probabilistic predictions.

Specifically, we rely on the sampling temperature $t$ of the VLM, where $t>0$ introduces randomness into the sampling procedure. Then, the main question becomes how to aggregate different predictions. As our main goal is to obtain probabilities for the mistake predictions, we first sample our action segmentation deterministically with $t=0$ to obtain the most likely segmentation $\hat{y}_{s,0}$, and its corresponding mistake labels $\hat{y}_{e,0}$. Then, we increase the temperature and generate $N$ more samples, $\{\hat{y}_{s,t,i}\}_{i=1}^N$ and $\{\hat{y}_{e,t,i}\}_{i=1}^N$. We match the segments of these $N$ predictions to the segment in the original segmentation in which their midpoint falls. Then, the mistake label for any frame is the sum of the error predictions for that segment divided by the total number of predictions (i.e., $N+1$).

%% file: sec/4_exps.tex
\section{Experiments}

\model~does not require any training samples, labels, or examples of errors, and is therefore generalizable to many tasks. We now proceed to the experimental results, where we show these benefits on real-world tasks.

\subsection{Experimental Set-up}

\textbf{Baselines.} As there are no previous attempt at zero-shot procedural mistake detection, direct comparisons to other methods are not possible. Instead, we include the state-of-the-art fully supervised approaches EgoPED~\citep{lee2024error}, AMNAR~\citep{huang2025modeling}, GTG2Vid~\citep{lee2025error}, and AEM~\citep{guo2025procedural} into our evaluation to provide a reference point. 

\noindent\textbf{Datasets.}
Following previous works (e.g., \cite{guo2025procedural}), we use two established and public benchmarks for our evaluation: EgoPER~\citep{lee2024error} and CaptainCook4D~\citep{peddi2024captaincook4d}. EgoPER contains egocentric videos spanning five recipes: pinwheels, quesadilla, oatmeal, coffee, and tea. There are 385 videos with a total duration of 28 hours, of which 213 videos are correct executions of the recipe, whereas the other 178 contain one or more mistakes. The mistakes come in five different types, namely omission, addition, modification, slip, and correction errors, and are defined per segment. We evaluate all models on the pre-defined test splits.
CaptainCook4D also consists of egocentric cooking videos. There are 384 videos, which are generally longer than EgoPER at a total duration of 94.5 hours, and a wider variety of 24 recipes. Of these videos, 164 are performed correctly, while 220 videos contain one or more errors. Following common practice, we consider all videos with errors as the test set. There are seven error categories: preparation, measurement, timing, temperature, missing step, technique, and order.

We separate these errors into segment-level and task-level errors. Segment-level errors are mistakes even when considering the segment in isolation, such as an action performed incorrectly. In contrast, task-level errors can only be seen in the context of the whole procedure, i.e., an omitted step.
Some previous works can only detect segment-level errors (e.g., \cite{guo2025procedural,huang2025modeling}, whereas others are capable of detecting both~\citep{lee2025error}. We split our evaluation explicitly into segment-level and task-level evaluations for clarity.

\noindent\textbf{Metrics.}
For segment-level errors, we compute the Error Detection Accuracy (EDA)~\citep{lee2024error,huang2025modeling} and the F1@.5, where 0.5 denotes the overlapping threshold between segments. We compute the F1@.5 by averaging over the F1 scores that are independently computed on the segments for each step or error type following~\cite{lee2025error}. For task-level errors, we follow \cite{lee2024error} and report the Omission Accuracy (O-Acc) and Omission Intersection over Union (O-IoU). We evaluate TAS by standard metrics: the Mean over Frames (MoF), IoU, Edit score, F1@K, and accuracy.

\noindent\textbf{Implementation details.}
We use all models with their default settings and set the frames-per-second to 4. For \model-MC, we set $t=0.7$. We ran the model on four H100 GPUs with 96GB of VRAM, 128 CPUs, and 500GB of RAM. We used VLLM version 0.19.1~\citep{kwon2023efficient} for hosting the VLMs~\citep{qwen3.5}, Python 3.12.3, and PyTorch version 2.10.0~\citep{paszke2019pytorch}.

\begin{table}[t]
    \caption{\textbf{Segment-level mistake detection performance on EgoPER.} We report the EDA and F1@.5. Results for baselines are obtained from their official checkpoints. For \model, we report results with Qwen3.5-35B-A3B (Q-M) and Qwen3.5-397B-A17B-GPTQ-Int4 (Q-L). Despite being zero-shot, \model~outperforms fully supervised methods on average in both metrics.} %
    \label{tab:egoper_segment}
    \centering
    \resizebox{\textwidth}{!}{
    \begin{tabular}{lccccccccccccc}
        \toprule
        \multirow{3}{*}{\textbf{Method}} & \multicolumn{12}{c}{\textbf{Recipe}} & \\  
        & \multicolumn{2}{c}{Quesadilla} & \multicolumn{2}{c}{Oatmeal} & \multicolumn{2}{c}{Pinwheel} & \multicolumn{2}{c}{Coffee} & \multicolumn{2}{c}{Tea} & \multicolumn{2}{c}{\textbf{Average}} \\
        \cmidrule(lr){2-3}\cmidrule(lr){4-5}\cmidrule(lr){6-7}\cmidrule(lr){8-9}\cmidrule(lr){10-11}
        & \textbf{EDA} & \textbf{F1@.5} & \textbf{EDA} & \textbf{F1@.5}
        & \textbf{EDA} & \textbf{F1@.5} & \textbf{EDA} & \textbf{F1@.5}
        & \textbf{EDA} & \textbf{F1@.5} & \textbf{EDA} & \textbf{F1@.5} \\  
        \midrule
        \rowcolor{white!80!pink} & \multicolumn{12}{c}{\textbf{Fully supervised}} & \\
        EgoPED & 62.7 & 27.7 &  51.4 & 21.1 & 59.6 & 19.6 & 55.3 & 9.2 & 56.0 & 35.1 & 57.0 & 22.5  \\
        AMNAR & 61.4 & 38.4 & 65.0 & 34.4 & 65.0 & 22.7  & 73.5 & 22.4  & 57.0 & 30.9 & 64.4 & 29.4  \\
        GTG2Vid & 79.9 & 31.5 & 86.5 & 47.4 & 70.8 & 23.3 & 86.1 & 18.6 & 75.4 & 44.2 & 79.7 & 33.0\\
        AEM & 68.1 &  45.2 & 68.6 & 27.0  & 61.2 & 41.8 & 66.4 & 39.3 & 69.4 &  41.5  & 66.7 & 39.0 \\
        \midrule
        \rowcolor{white!80!pink} & \multicolumn{12}{c}{\textbf{Zero-shot}} & \\
        \textbf{\model} (Q-M) & 82.0 & 46.2 & 85.3 & 31.5 & 77.8 & 35.1 & 76.5 & 28.6 & 79.0 & 32.5 & 80.1 & 34.8  \\
        \textbf{\model} (Q-L) & 85.2 & 51.5 & 83.1 & 34.0 & 80.2 & 36.3 & 89.3 & 41.2 & 82.7 & 41.9 & 84.1 & 41.0 \\
        \bottomrule
    \end{tabular}
    }
\end{table}

\begin{table}[t]
    \caption{\textbf{Segment-level mistake detection performance on CaptainCook4D.} We compare to reported numbers of EDA and F1@.5, for which GTG2Vid only uses a subset of 5 recipes instead of the full dataset. \model~reaches a performance close to fully supervised methods.} %
    \label{tab:cc4d_segment}
    \centering
    \resizebox{0.8\textwidth}{!}{
    \begin{tabular}{lcccc}
        \toprule
        \multirow{3}{*}{\textbf{Method}} & \multicolumn{2}{c}{\multirow{2}{*}{\textbf{Average} }} & \multicolumn{2}{c}{\multirow{2}{*}{\textbf{5-recipe average} }} \\
        & \multicolumn{2}{c}{} & \multicolumn{2}{c}{} \\
        \cmidrule(lr){2-3} \cmidrule(lr){4-5}
        & \textbf{EDA} & \textbf{F1@.5} & \textbf{EDA} & \textbf{F1@.5} \\  
        \midrule
        \rowcolor{white!80!pink} &\multicolumn{4}{c}{\textbf{Fully supervised}} \\
        EgoPED~\citep{lee2024error} &  69.8 &  - & - & 10.5 \\
        AMNAR~\citep{huang2025modeling} & 72.3 & -  & - & -\\
        GTG2Vid~\citep{lee2025error}  & - & - & 66.3 & 16.2  \\
        AEM~\citep{guo2025procedural} & 71.9 &  - & - & -\\
        \midrule
         \rowcolor{white!80!pink}&\multicolumn{4}{c}{\textbf{Zero-shot}}   \\
        \textbf{\model} (Q-M) & 69.6 & 14.0 & 66.3 & 10.2 \\
        \textbf{\model} (Q-L) & 69.0 & 17.9 & 67.4 & 16.0 \\
        \bottomrule
    \end{tabular}
    }
    \vspace{-6mm}
\end{table}

\subsection{Main results}
\noindent\textbf{Segment-level errors.}
We show our main results on segment-level errors of EgoPER in Tab.~\ref{tab:egoper_segment} and CaptainCook4D in Tab.~\ref{tab:cc4d_segment}, where we find that \model~achieves promising results across all metrics.
As \model~is the first zero-shot framework for procedural mistake detection and no direct comparison to previous methods is possible, we additionally report fully supervised methods to provide a frame of reference.
The results on EgoPER show that, out of the fully supervised methods, GTG2Vid consistently outperforms the other supervised methods in EDA, leading to an average EDA of 79.7 across the five recipes. In contrast, AEM scores the highest F1@.5 of all supervised methods. Yet, despite being zero-shot, \model~outperforms both GTG2Vid and AEM in both metrics, reaching an average EDA of 84.1, improving the results on all recipes except oatmeal, and an average F1@.5 of 41.0. 
For CaptainCook4D, we reach slightly lower performance than EgoPED, AMNAR, and AEM on the entire dataset, for example, trailing the state-of-the-art method in EDA by $2.7$ points.
GTG2Vid is evaluated on a different subset than the other models, using only five recipes out of a total of 24. For completeness, we ran \model~also on this subset. Similar to the full dataset, \model~only slightly falls behind the fully supervised GTG2Vid, coming within $0.2$ points in F1@.5.

We want to stress that these performance gains are not simply due to the use of large models: baselines such as AEM and GTG2Vid also make use of LLMs and VLMs as building blocks in their approach. For instance, AEM uses GPT-4o to generate scene graphs from individual frames. However, by tackling the full task with the VLM, rather than only using it as a building block, \model~matches or even outperforms these methods in many cases.

\begin{table}[t]
    \caption{\textbf{Task-level mistake detection performance on EgoPER.} We report the O-Acc and O-IoU for methods capable of detecting omission errors. Results for baselines are obtained from their official checkpoints. \model~approaches supervised-level performance without training.} 
    \label{tab:egoper_task}
    \centering
    \resizebox{\textwidth}{!}{
        \begin{tabular}{lcccccccccccc}
        \toprule
        \multirow{3}{*}{\textbf{Method}} & \multicolumn{11}{c}{\textbf{Recipe}} & \\ 
        & \multicolumn{2}{c}{Quesadilla} & \multicolumn{2}{c}{Oatmeal} & \multicolumn{2}{c}{Pinwheel} & \multicolumn{2}{c}{Coffee} & \multicolumn{2}{c}{Tea} & \multicolumn{2}{c}{\textbf{Average}} \\
        \cmidrule(lr){2-3}\cmidrule(lr){4-5}\cmidrule(lr){6-7}\cmidrule(lr){8-9}\cmidrule(lr){10-11}
        & \textbf{O-Acc} & \textbf{O-IoU} & \textbf{O-Acc} & \textbf{O-IoU}
        & \textbf{O-Acc} & \textbf{O-IoU} & \textbf{O-Acc} & \textbf{O-IoU}
        & \textbf{O-Acc} & \textbf{O-IoU} & \textbf{O-Acc} & \textbf{O-IoU} \\  
        \midrule
        \rowcolor{white!80!pink} & \multicolumn{11}{c}{\textbf{Fully supervised}} & \\
        EgoPED &  77.1 & 66.1 & 82.5 &  72.2 & 63.2 & 54.0 & 47.4 & 39.1 & 55.3 & 39.4 & 65.1 & 54.2 \\
        GTG2Vid & 64.6 & 50.8 &  95.2 & 70.6 & 76.1 & 47.8 & 52.6 & 30.3 & 87.2 & 42.7 & 75.1 & 48.4 \\
        \midrule
        \rowcolor{white!80!pink} &\multicolumn{11}{c}{\textbf{Zero-shot}} & \\
        \textbf{\model} (Q-M) & 79.2 & 48.1 & 96.8 & 46.9 & 66.1 & 33.3 & 47.4 & 8.33 & 44.7 & 17.5  & 66.8 & 30.8 \\
        \textbf{\model} (Q-L) & 83.3 & 52.6 & 93.6 & 80.6 & 65.2 & 46.6 & 52.6 & 10.5 & 66.0 & 27.0 & 72.1 & 43.5\\
        \bottomrule
    \end{tabular}
    }
    \vspace{-5mm}
\end{table}

\noindent\textbf{Task-level errors.}
We now turn to task-level mistakes, specifically, omissions of certain steps in a procedure. We consider EgoPER, as no other method provides results or checkpoints for this task on CaptainCook4D. 
Detecting these task-level errors requires understanding the entire video. As such, segment-level only methods such as AMNAR and AEM are excluded from our comparison in Tab.~\ref{tab:egoper_task}, which shows that \model~is also successful in detecting omission errors. Compared to the fully-supervised methods, \model~outperforms EgoPED in O-Acc, but trails GTG2Vid by $3.0$ O-Acc and $4.9$ O-IoU. Nonetheless, the fact that these complicated errors are detected zero-shot with high accuracy shows the holistic video understanding capabilities of \model.

\noindent\textbf{Temporal action segmentation.}
Additionally, we report the TAS results on the five recipes of EgoPER in Tab.~\ref{tab:seg}. 
\model~is able to produce high-quality segmentations of the procedural videos.
Compared to the fully supervised methods, we find that there is no single best method, with a different method scoring highest in each of the four metrics.

\subsection{Qualitative results and failure cases}

\begin{figure}
    \centering
    \includegraphics[width=0.99\linewidth]{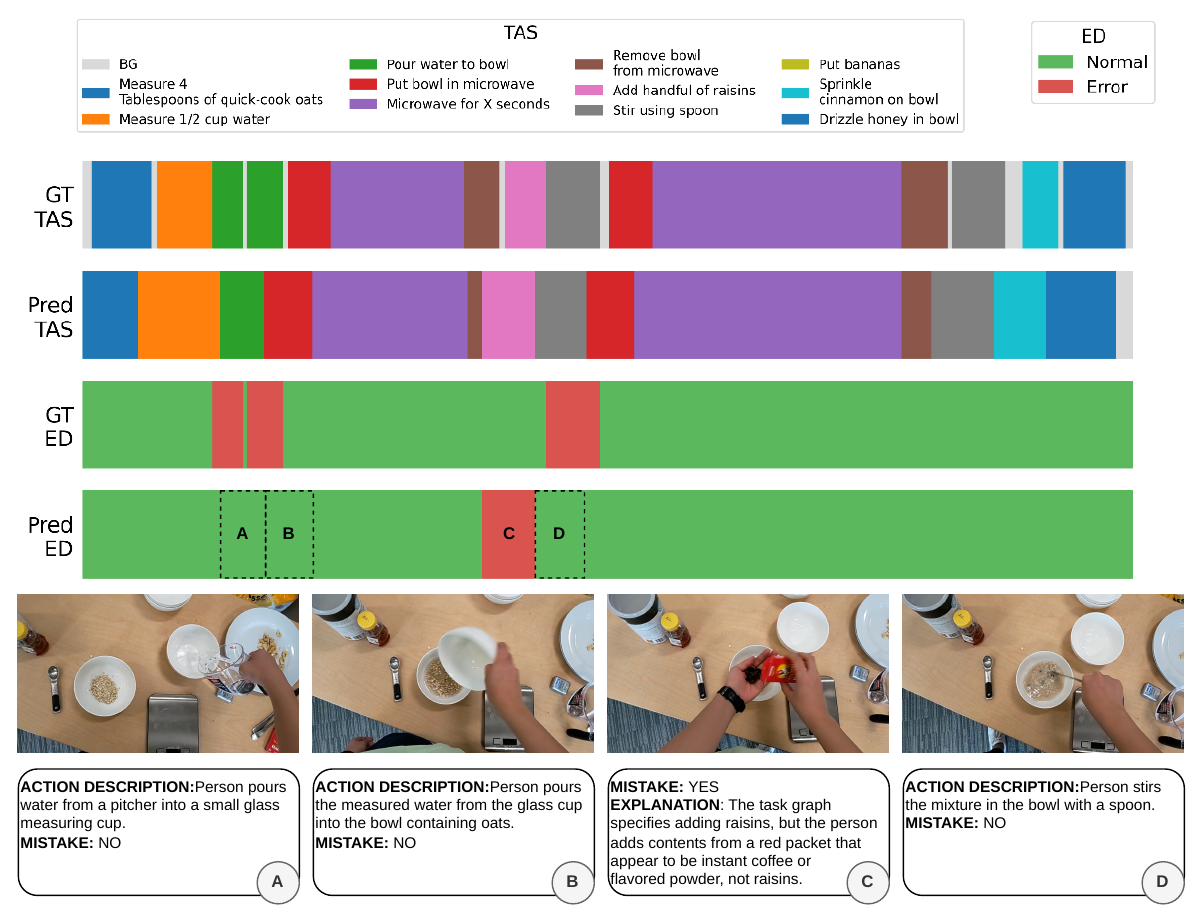}
    \caption{\textbf{Limitations of dataset labels and \model~on EgoPER.} 
    The GT labels the two ``pour water into bowl" segments as mistakes, because the person first pours water into a cup, then into the bowl. \model~considers this a single correct segment pouring water into the bowl. Together with ambiguous start and end frames, these label ambiguities negatively affect \model's performance, yet are unrelated to model quality.
    The other mistakes are due to \model~itself: in segment C, it confuses the raisins for coffee, while in segment D, it considers the knife a spoon.}
    \label{fig:expl}
    \vspace{-1mm}
\end{figure}

While the performance of \model~is, in general, surprisingly high and can compete with fully supervised methods, there are still some interesting failure cases. 
Fig.~\ref{fig:expl} shows such an example. 
One interesting quirk of zero-shot methods is that the arbitrary nature of segment labels leads to predictions that are evaluated as wrong, even though they are correct in another, equally valid, interpretation. For instance, the two ``pour water into bowl" segments are labeled as mistakes, because the person first pours water into a cup, then into the bowl. \model~considers this a single segment, where the water is correctly poured into the bowl. In general, we find that the performance of \model~is relatively low on these ambiguous correction errors, as we show in the appendix, and whether corrections should be considered as mistakes is debatable in the first place.
Moreover, the exact start and end frames of the segment are not well-defined, as there is empty space between actions, and multiple frames could denote the boundary. Both of these effects negatively affect \model's performance, yet stem from factors unrelated to model quality. Therefore, we believe the \model's zero-shot performance to be a lower bound, limited by the label ambiguity on which supervised methods overfit. Other mistakes come from limitations of \model~itself: segments C and D are both due to \model~hallucinating or misunderstanding the visual content.

\begin{figure}[t]
  \centering

  {\color{actiocolor}\textbf{Predicted action: Put coffee beans in the coffee grinder}}

  \vspace{3pt}
  \framerow{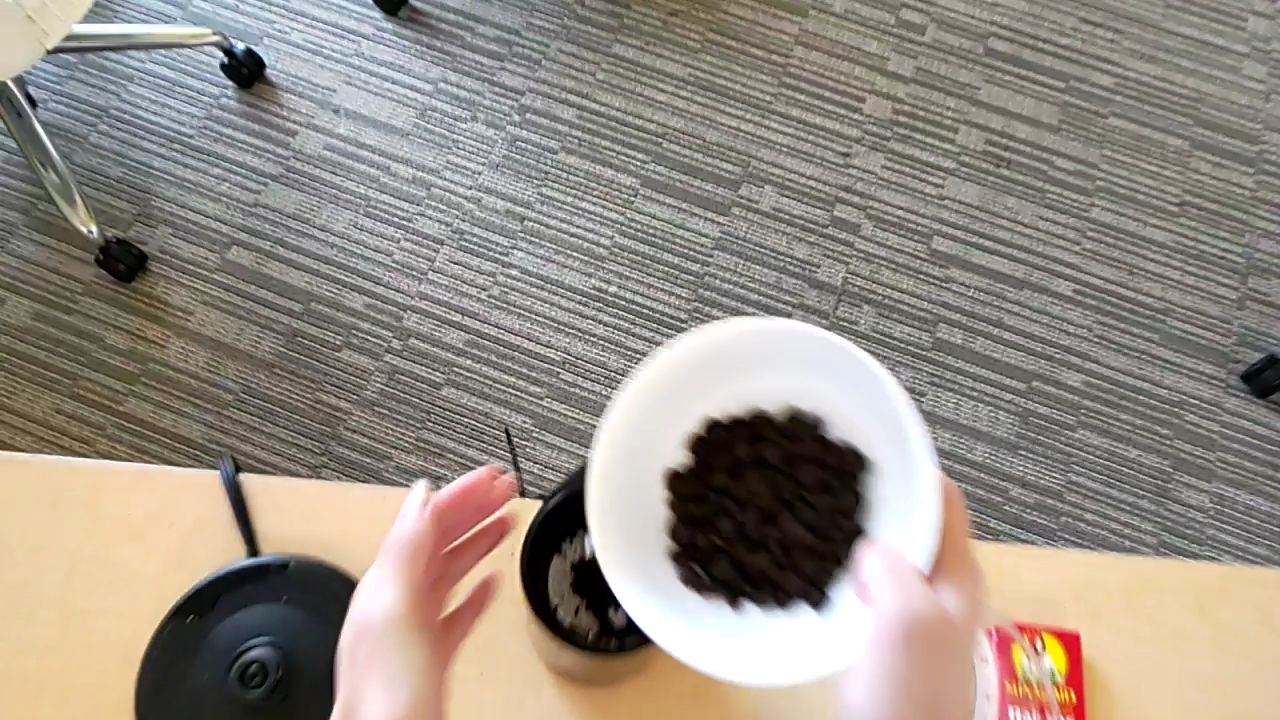}{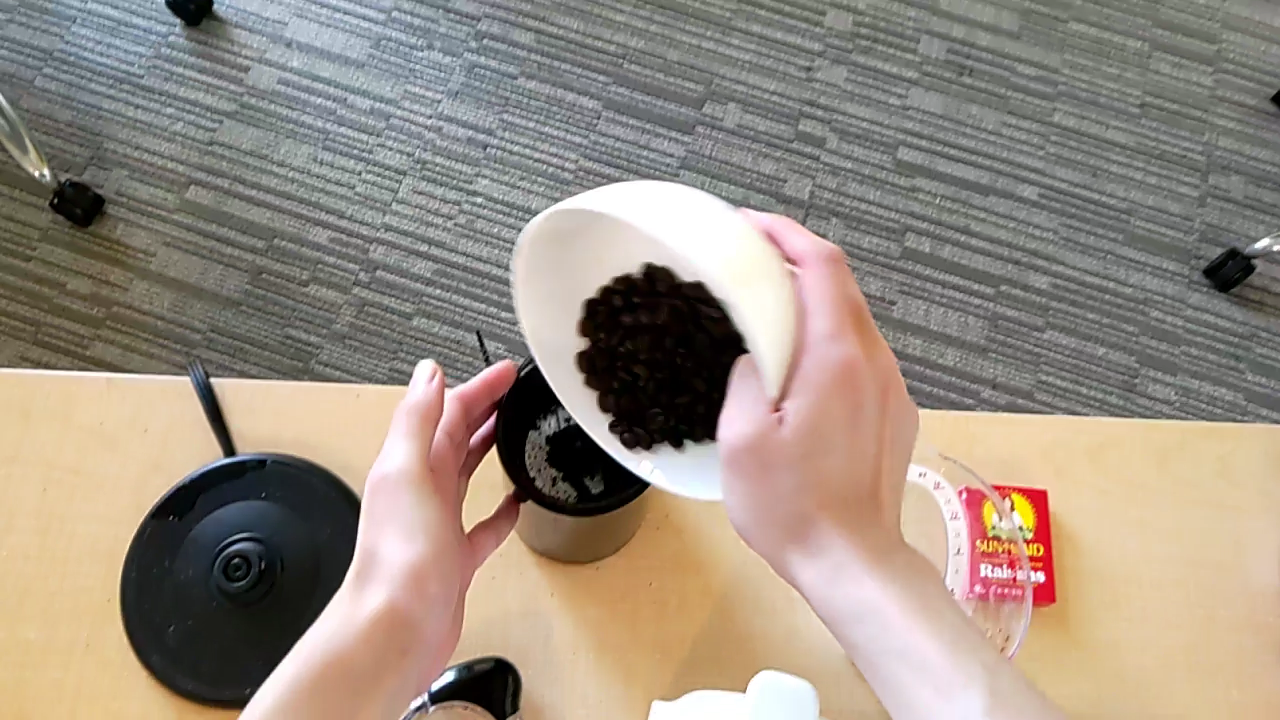}{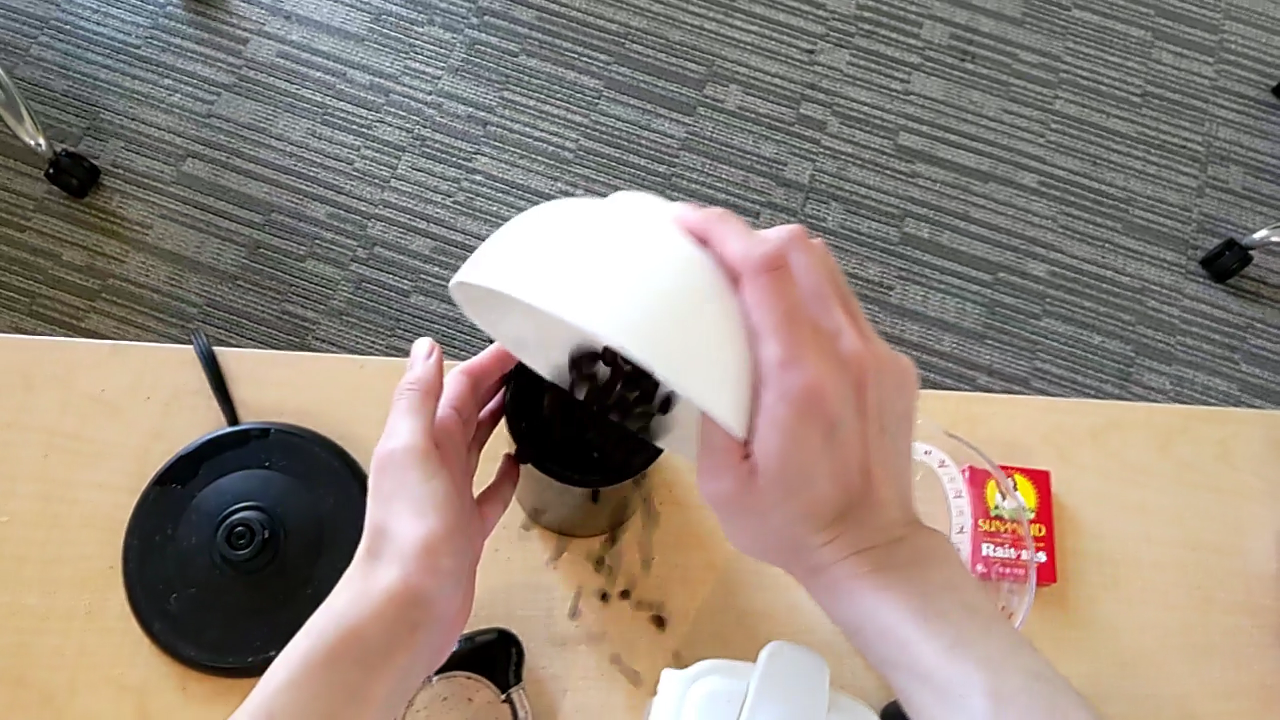}{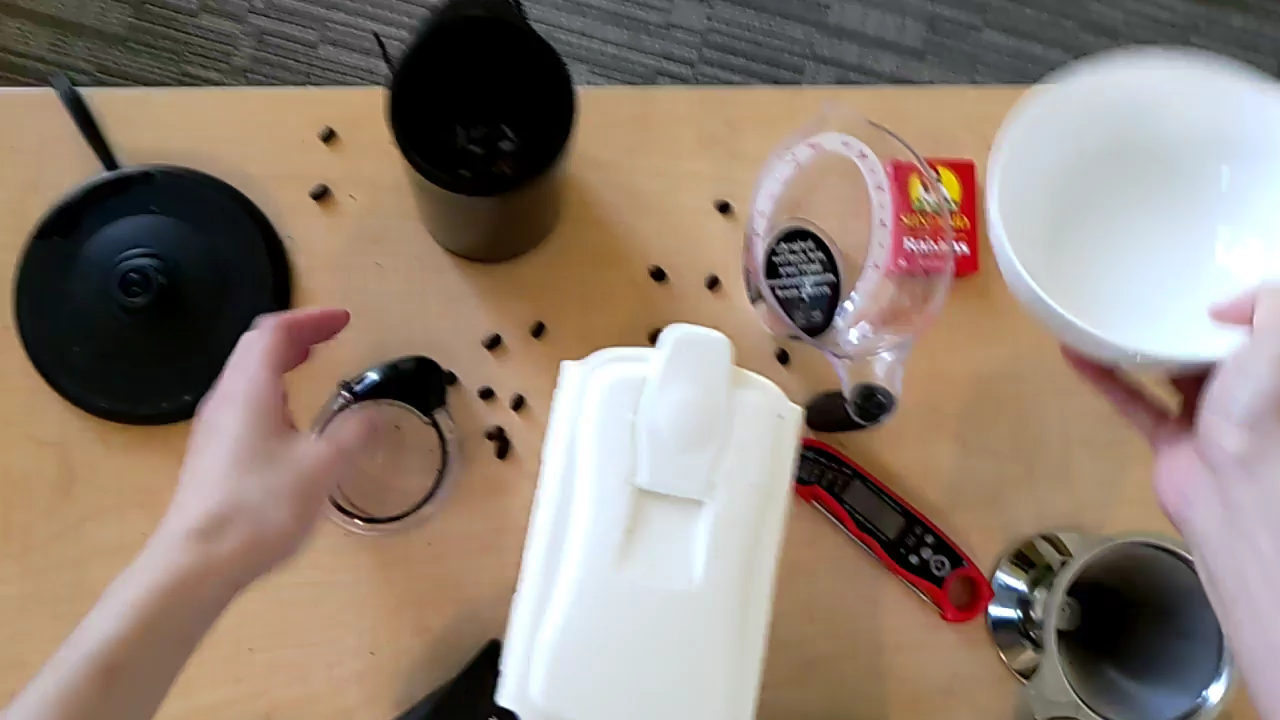}{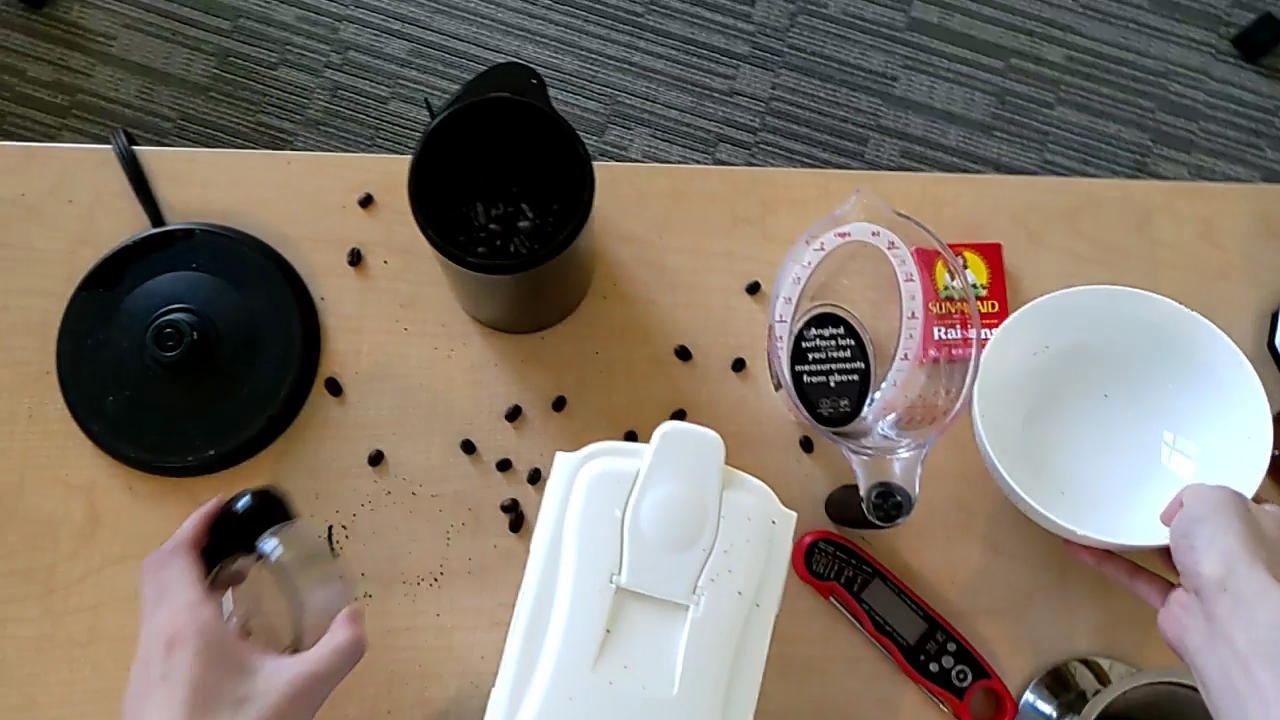}

  \vspace{4pt}
  \begin{subfigure}{\linewidth}
  \captionsetup{labelformat=empty}
    \caption{\textbf{Action description by \model}: "Person attempts to pour weighed coffee beans from the bowl into the grinder jar, spilling many beans onto the table." \\\textbf{Error explanation by \model}: "Although the intent matches, significant spillage occurred during transfer, indicating improper execution requiring cleanup later."}
  \end{subfigure}

  \caption{%
    \textbf{Qualitative example of explainable mistake detection on EgoPER's coffee recipe.} The frames show an example of a ``slip" error, where the coffee beans are dropped on the table rather than put into the grinder. \model~matches the segment to the correct action, describes what is shown in the segment, then correctly explains why this is an error compared to its predicted step.
  }
  \label{fig:mistakes}
\end{figure}

Besides segmenting and detecting mistakes, \model~provides explainable descriptions as to why a segment is labeled as an error. We show an example of this in Fig.~\ref{fig:mistakes}, where a mistake segment is correctly detected, described, and explained.

\begin{table}[tbp]
    \begin{center}
    \caption{\textbf{Temporal action segmentation performance on EgoPER and 50Salads.} We report the average performance over the five recipes for EgoPER, and over the standard five cross-validation splits for 50Salads. Higher is better for all metrics. \model~produces high-quality temporal action segmentations. We use Qwen3.5-397B-A17B-GPTQ-Int4 for \model.}
    \begin{minipage}{0.48\textwidth}    
        \centering
        \subcaption{EgoPER}
        \label{tab:seg}
        \resizebox{0.85\textwidth}{!}{
            \begin{tabular}{lcccc}
            \toprule
            Method & IoU & Edit & F1@0.5 & Acc \\
            \midrule
            \rowcolor{white!80!pink} &\multicolumn{3}{c}{\textbf{Fully supervised}} & \\
            EgoPED  & 44.6 & 61.3 & 47.5 & 68.5 \\
            AMNAR   & 56.3 & 69.4 & 57.3 & 75.3 \\
            GTG2Vid & 58.0 & 75.4 & 70.4 & 70.3 \\
            AEM & 58.5 & 69.7 & 58.5 & 73.5 \\
            \midrule
            \rowcolor{white!80!pink} &\multicolumn{3}{c}{\textbf{Zero-shot}} & \\
            \textbf{\model} & 48.8 & 81.0 & 59.5 & 64.1 \\
            \bottomrule
            \end{tabular}
        }
    \end{minipage}    
    \hfill    
    \begin{minipage}{0.48\textwidth}    
        \centering
        \subcaption{50Salads}
        \label{tab:tas}
        \resizebox{\textwidth}{!}{ 
            \begin{tabular}{lccccc}
                \toprule
                Method & MoF & Edit & F1@10 & F1@25 & F1@50 \\
                \midrule
                \rowcolor{white!80!pink} &\multicolumn{4}{c}{\textbf{Fully supervised}} & \\
                MS-TCN & 80.7 & 67.9 & 76.3 & 74.0 & 64.5 \\
                LTC & 87.7 & 83.2 & 89.4 & 87.7 & 82.0 \\
                DiffAct & 88.9 & 85.0 & 90.1 & 89.2 & 83.7 \\
                \midrule
                \rowcolor{white!80!pink} &\multicolumn{4}{c}{\textbf{Zero-shot}} & \\
                OVTAS & 31.5 & \textbf{88.7} & 42.6 & 31.4 & 14.1 \\
                \textbf{\model} & \textbf{68.9} & 67.8 & \textbf{70.0} & \textbf{66.0} & \textbf{56.0} \\
                \bottomrule
            \end{tabular}
        }    
    \end{minipage}   
    
    \end{center}
\end{table}

\subsection{Temporal action segmentation}
Before in Tab.~\ref{tab:seg}, we compared our TAS results on procedural mistake detection benchmarks against the TAS modules of mistake detection methods. Here, we additionally provide experiments on a standard TAS benchmark, 50 Salads~\citep{stein2013combining}, following the standard 5-split cross-validation of \cite{farha2019ms}.
From Tab.~\ref{tab:tas}, we find that \model~outperforms the only previous zero-shot TAS method, OVTAS~\citep{unmesh2026exploring}. For instance, it improves upon OVTAS by $41.9$ points in F1@50. While it still falls behind specialized supervised methods on this task (MS-TCN~\citep{farha2019ms}, LTC~\citep{bahrami2023much}, and DiffAct~\citep{liu2023diffusion}), the segmentations of \model~are surprisingly high-quality, as also shown in the qualitative examples.

\begin{table}[t]
    \caption{\textbf{Probabilistic predictions on segment-level mistake detection for EgoPER.} We report the EDA, AUC, and F1@.5 with Qwen3.5-35B-A3B. Using more samples leads to improved AUC scores.} 
    \label{tab:egoper_segment_prob}
    \centering
    \resizebox{\textwidth}{!}{
    \begin{tabular}{lccccccccccccc}
        \toprule
        \multirow{3}{*}{\textbf{Method (Q-L)}} & \multicolumn{11}{c}{\textbf{Recipe}} & \\ 
        & \multicolumn{2}{c}{Quesadilla} & \multicolumn{2}{c}{Oatmeal} & \multicolumn{2}{c}{Pinwheel} & \multicolumn{2}{c}{Coffee} & \multicolumn{2}{c}{Tea} & \multicolumn{2}{c}{\textbf{Average}} \\
        \cmidrule(lr){2-3}\cmidrule(lr){4-5}\cmidrule(lr){6-7}\cmidrule(lr){8-9}\cmidrule(lr){10-11}
        & \textbf{EDA} & \textbf{AUC} & \textbf{EDA} & \textbf{AUC}
        & \textbf{EDA} & \textbf{AUC} & \textbf{EDA} & \textbf{AUC}
        & \textbf{EDA} & \textbf{AUC} & \textbf{EDA} & \textbf{AUC} \\  
        \midrule
        \rowcolor{white!80!pink} &\multicolumn{11}{c}{\textbf{Zero-shot}} & \\
        \textbf{\model-MC (n=1)} & 85.2 & 67.7 & 83.1 & 53.7 & 80.2 & 62.4 & 89.3 & 57.0 & 82.7 & 63.4 & 84.1 & 60.8 \\
        \textbf{\model-MC (n=5)} & 84.5 & 70.3 & 88.9 & 58.2 & 82.8 & 73.8 & 92.2 & 56.5 & 85.1 & 68.9 & 86.7 & 65.5 \\
        \textbf{\model-MC (n=10)} & 83.5 & 72.1 & 89.2 & 58.3 & 81.3 & 77.2 & 92.3 & 55.3 & 84.2 & 71.8 & 86.1 & 66.9 \\
        \bottomrule
    \end{tabular}
    }
\end{table}

\subsection{Evaluating probabilistic predictions}
While our base \model~only assigns binary error labels to segments, \model-MC assigns error probabilities. As a result, we can evaluate \model-MC with metrics that assume probabilistic predictions that need to be thresholded. In particular, the AUC is commonly used to evaluate the binary mistake detection task~\citep{guo2025procedural}, and we evaluate the effect of the number of samples on this metric to assess whether the probabilities obtained by \model-MC improve with more samples.

Tab.~\ref{tab:egoper_segment_prob} shows that increasing the number of samples from one to five improves the EDA by $2.6$, and especially the AUC by $4.7$ points. Increasing the number of samples further to 10 leads to a small decrease in EDA but a further increase in the AUC of $1.4$ points.
We note that this does come at the cost of $N$ inference calls to a large VLM. Therefore, while higher number of samples leads to increased performance, when to use \model-MC, and with how many samples, is use-case specific, and we stick to \model~in our main experiments.

\subsection{Ablation studies}

\begin{table}[t]
    \begin{center}
    \caption{\textbf{Ablation studies on EgoPER.} We use Qwen3.5-397B-A17B-GPTQ-Int4 for all experiments. (left) We ablate the effect of the procedure input representation on the performance. Using an ordered list of actions leads to the best performance. (right) Ablation of the output representation. We find that jointly solving TAS and mistake detection leads to the best performance. }
    \begin{tabular}{lcc}
        \toprule
        Actions Input & EDA & F1 \\
        \midrule
        DOT-graph  & 81.4 & 40.2 \\
        DAG & 82.5 & 40.9 \\
        \textbf{\model} &  \textbf{84.1} & \textbf{41.0} \\
        \bottomrule
        \end{tabular}
        \quad\quad
        \begin{tabular}{lcc}
        \toprule
        Prediction Output & EDA & F1  \\
        \midrule
        No-TAS &  79.2 & 32.8 \\
         \textbf{\model} &  \textbf{84.1} & \textbf{41.0} \\
        \bottomrule
    \end{tabular}
    \label{tab:abl}
    \end{center}
\end{table}

\noindent\textbf{Input format.} We provide the procedure instructions as an ordered list to the VLM. In Tab.~\ref{tab:abl} (left), we show that this outperforms the procedure as a directed acyclic graph (DAG) or as a DOT-graph.

\noindent\textbf{Output format.}
% We instruct the VLM to output \texttt{JSON} files with a specific format. Here, we show that our design of this structure has a large effect on the performance. 
We compare to a No-TAS approach, where the model performs error detection directly rather than providing the temporal segmentation as well. Tab.~\ref{tab:abl} (right) shows that jointly solving TAS and mistake detection leads to the best performance.

\subsection{Limitations}\label{sec:limit}
We rely on large VLMs to do our zero-shot procedural mistake detection, and full models can be large and difficult to run on consumer-grade hardware. However, memory optimization techniques such as quantization make their usage feasible even in constrained settings. Our main results with the Qwen3.5-397B-A17B-GPTQ-Int4 model show that, despite aggressive quantization, the performance \model~remains high and can be run on a single node with 4 H100 GPUs. On this node, processing the full EgoPER dataset takes 87.8 minutes.
Still, these models are large and, as a result, \model~requires modern GPUs to use. 
Moreover, as we show in our qualitative results, even though \model~approximates fully supervised results in many cases, there are still some segments and mistakes misdetected due to hallucinations or limitations of labels in existing datasets, leaving substantial room for improvement. 

%% file: sec/6_conclusion.tex
\section{Conclusion}

In this work, we introduced zero-shot procedural mistake detection and proposed \model.
We showed that \model~enables state-of-the-art VLMs to simultaneously perform action segmentation and mistake detection in a zero-shot fashion by using them as a standalone module rather than a component in a larger pipeline. On both EgoPER and CaptainCook4D, \model~can approach or even outperform fully supervised methods, without needing any task-specific training, data, or annotations. As a result, our framework is readily applicable to unseen tasks. Future work will explore the benefits of lightweight adaptation, for instance, via prompt learning~\citep{miyai2023locoop}, to increase performance further at the cost of some computational resources.
Overall, our results show the potential of unified methods for procedural mistake detection, and we hope this will steer the field away from highly complex pipelines and toward more generally applicable solutions.

%% file: sec/7_appendix.tex
\newpage

\section*{Appendix}

\section{Per-mistake performance}
\begin{table}[h]
\caption{\textbf{Performance per error type.} We show results with Qwen3.5-397B-A17B-GPTQ-Int4 (Q-L) on EgoPER. \model~performs relatively worse on correction errors due to their ambiguous nature, but detects all other mistakes at a similar rate.}
\centering
\begin{tabular}{lcccc}
\toprule
\textbf{Task} & Modification & Slip & Correction & Addition \\
\midrule
Quesadilla & 59.7 & 56.0 &  58.4 & 72.5 \\
Oatmeal    & 37.6 & 35.4 &  29.7 & 14.6 \\
Pinwheel   & 45.9 & 31.9 &  17.9 & 44.6 \\
Coffee     & 10.2 & 40.8 & 28.6 &  -  \\
Tea        & 60.2 & 57.2 &  24.6 & 59.9 \\
\midrule
\textbf{Average} & 42.7 & 44.3 & 31.8 & 47.9 \\
\bottomrule
\end{tabular}

\label{tab:egoper_breakdown}
\end{table}

We show the performance per segment-level mistake type (addition, modification, slip, and correction) on EgoPER in Tab.~\ref{tab:egoper_breakdown}. Overall, we find that the performance of \model~is relatively low on correction errors. As discussed in the main paper, this comes from ambiguous labeling, and whether corrections should be considered as mistakes is debatable in the first place. The performance is similar on the other mistake types, between $42.7$ F1@.5 for modifications and $47.9$ F1@.5 for additions.

\section{Additional Qualitative Results}\label{sec:app_qual}

\begin{figure}[h]
    \centering
    \includegraphics[width=0.95\linewidth]{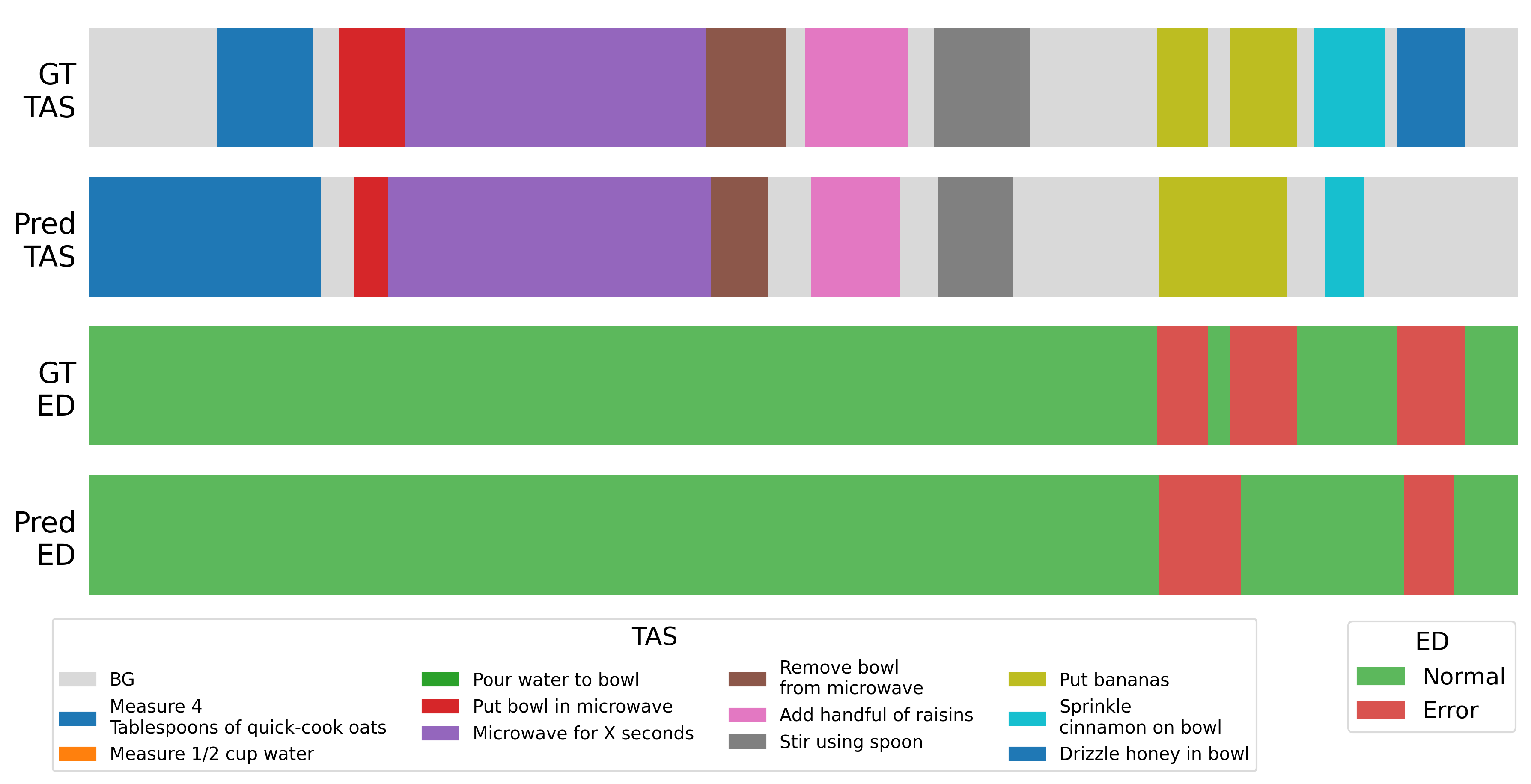}
    \caption{\textbf{Zero-shot procedural mistake detection with \model.} This result belongs to the ``oatmeal" task from EgoPER.}
    \label{fig:app_oatmeal}
\end{figure}

\begin{figure}[h]
    \centering
    \includegraphics[width=0.95\linewidth]{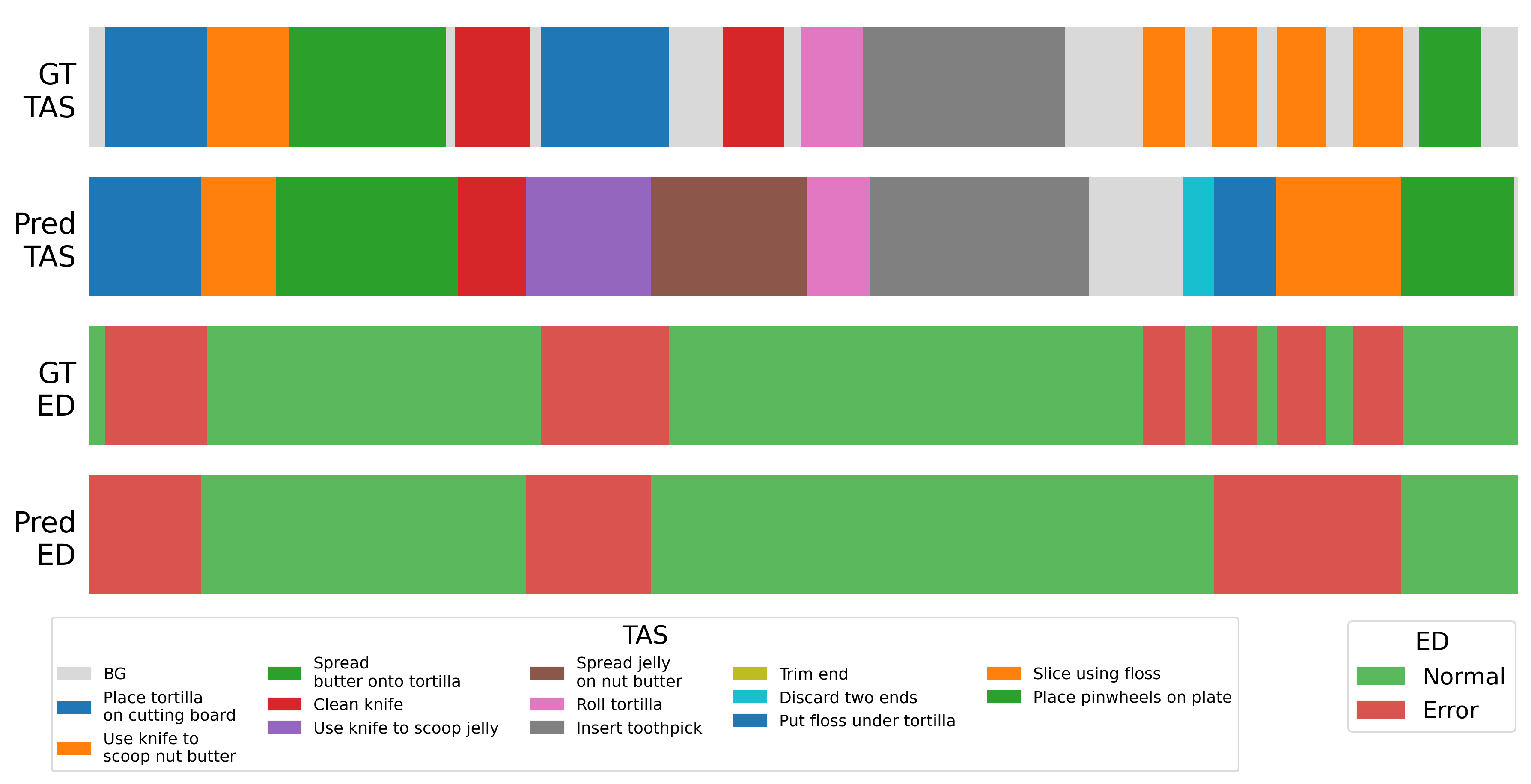}
    \caption{\textbf{Zero-shot procedural mistake detection with \model.} This result belongs to the ``pinwheels" task from EgoPER.}
    \label{fig:app_pinwheels}
\end{figure}

\begin{figure}[h]
    \centering
    \includegraphics[width=0.95\linewidth]{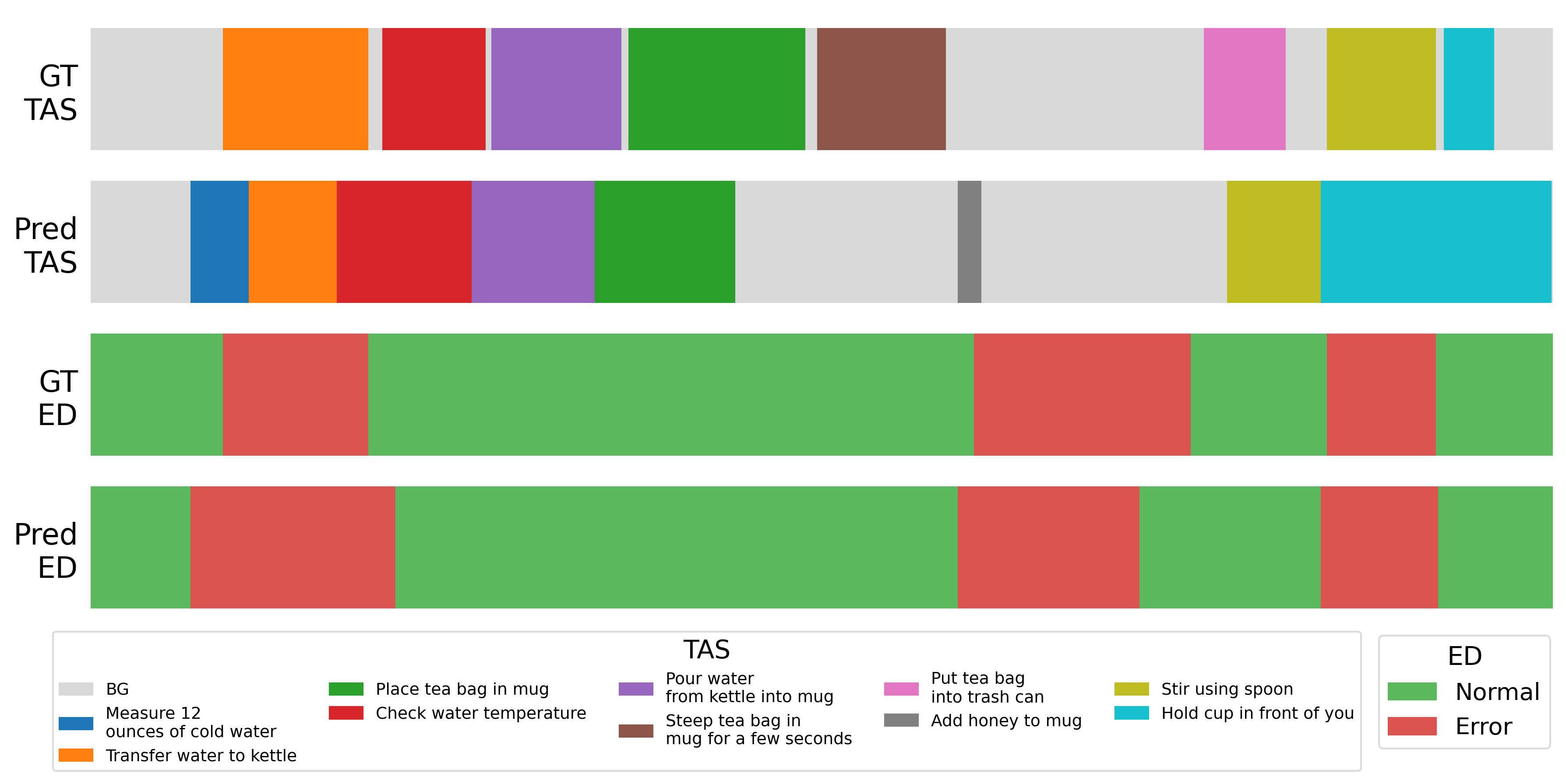}
    \caption{\textbf{Zero-shot procedural mistake detection with \model.} This result belongs to the ``tea" task from EgoPER.}
    \label{fig:app_tea}
\end{figure}

In this section, we show further qualitative results of our model for different tasks of EgoPER datasets. While Fig.~\ref{fig:app_oatmeal} belongs to the oatmeal task, Fig.~\ref{fig:app_pinwheels} and Fig.~\ref{fig:app_tea} are samples from the pinwheels and tea tasks, respectively. 

\section{Full prompt}
\begin{promptbox}{VLM Prompt Template}
    \small
    You are an expert at analyzing cooking procedure videos and detecting mistakes.

    \texttt{\{task\_graph\_block\}}
    
    Watch this video carefully and perform the following tasks:

    \begin{enumerate}
        \item SEGMENT the video into discrete action segments. For each segment provide:
        \begin{itemize}
            \item[-] start and end timestamps (in seconds)
            \item[-] a concise description of the action performed
            \item[-] Total duration of all segments MUST equal the exact video length (final end\_time\_sec = last frame timestamp). Verify this before finalizing. No gaps or overlaps allowed.
            \item[-] Typical cooking actions last 5-30 seconds. If a segment would exceed 45 seconds, re-examine it: it very likely contains multiple distinct actions that should be split.
            \item[-] When uncertain whether to split, prefer splitting over merging.
            \item[-] Start a new segment whenever: the person picks up or puts down an object, changes the object they are interacting with, or begins a visually distinct motion (e.g., switching from spreading to sprinkling).
            \item[-] End a segment at the last moment of active motion. Do NOT let idle pauses or brief hesitations extend a segment's end time.
        \end{itemize}
        
        \item For each segment, MATCH it to the task graph steps above using these rules:
        \begin{itemize}
            \item[-] The matched\_step must be the exact step name from the task graph (or "background" /"unexpected").
            \item[-] Compare the observed action precisely against every key detail of the matched step: location (e.g., "cutting board" vs "table"), object (e.g., "Nutella" vs "butter"), and tool (e.g., "knife" vs "spoon").
            \item[-] If all key details match -> has\_error = false.
            \item[-] If any key detail differs (even though the general action type is right) -> has\_error = true, error\_type = "Wrong execution", and explain the specific mismatch in error\_explanation.
        \end{itemize}

        \item DETECT ERRORS using these precise rules:
        \begin{enumerate}
            \item Wrong execution: the step matches a task graph step in type but not in key detail (e.g., tortilla placed on table instead of cutting board). \\
            -> has\_error = true, error\_type = "Wrong execution"
            \item Wrong order / missing prerequisite: a step is performed before its prerequisites are completed. \\
            -> Do NOT flag the step itself as having an error. The physical action may be correct.
            Instead, add the unfinished prerequisite steps to missing\_steps. \\
            -> The segment where the out-of-order step is performed should have has\_error = false
            as long as the action itself was correctly executed.
            \item Missing step: a required step was never performed at any point in the video. \\
            -> Record it in missing\_steps. Do not propagate this as an error on later steps.
            \item Wrong action: an action is performed that has no correspondence in the task graph. \\
            -> has\_error = true, error\_type = "Wrong action", matched\_step = "unexpected"
        \end{enumerate}

        \item BACKGROUND classification rules:
        \begin{itemize}
            \item[-] Only classify a segment as "background" (matched\_step = "background") if it involves NONE of the main task objects: tortilla, knife, Nutella, banana, cinnamon, plate, cutting board.
            \item[-] If the activity touches or affects any main task object, do NOT classify it as background. Instead, treat it as a proper task segment and evaluate it for errors.
        \end{itemize}

        \item List any task graph steps that were NEVER performed (missing\_steps).    

        \item Give an OVERALL VERDICT: "correct" if no errors were found, "has\_mistakes" if any has\_error = true or missing\_steps is non-empty.

    \end{enumerate}
    Be precise with timestamps. If a step is missing entirely, still flag it in missing\_steps.
\end{promptbox}

The \texttt{task\_graph\_block}, for the example of the EgoPER recipe ``oatmeal", looks as follows:

\begin{promptbox}{TASK GRAPH}
     % (action segmentation with indices):
  Action segments:
  
    [0] background
    
    [1] Measure 4 Tablespoons of quick-cook oats
    
    [2] Measure 1/2 cup water
    
    [3] Pour water to the bowl
    
    [4] Put bowl in the microwave
    
    [5] Microwave for X seconds
    
    [6] Remove the bowl from the microwave
    
    [7] Add handful of raisins
    
    [8] Stir using spoon
    
    [9] Sprinkle cinnamon on the bowl
    
    [10] Drizzle honey in the bowl
    
\end{promptbox}

\section{Variability in results}
While \model~is a deterministic method, we can increase the sampling temperature to obtain randomness between individual runs and assess the variance in the results. We do so for the ``quesadilla" recipe of EgoPER and report the results in Tab.~\ref{tab:var}, which shows that the variation in performance is small.

\begin{table}[b]
    \caption{\textbf{Standard deviation in segment-level mistake detection performance for \model.} We report standard deviation for EDA and F1@.5 with Qwen3.5-397B-A17B-GPTQ-Int4 (Q-L) when using $t=0.7$ over five runs. } 
    \label{tab:var}
    \centering
    \begin{tabular}{lccc}
        \toprule
         & \multicolumn{2}{c}{\textbf{Recipe}} & \\ 
        & \multicolumn{2}{c}{Quesadilla}  \\
        \cmidrule(lr){2-3}
        & \textbf{EDA} & \textbf{F1@.5}\\  
        \midrule
        \rowcolor{white!80!pink} & \multicolumn{2}{c}{\textbf{Standard deviation}} & \\
        \textbf{\model} (Q-L) & 0.4 & 1.7 \\
        \bottomrule
    \end{tabular}
    % }
\end{table}

\section{Licenses}
List of licenses for existing assets used in this paper.

\textbf{Models:}
\begin{itemize}
    \item Qwen3.5-35B-A3B~\citep{qwen3.5}: Apache license 2.0
    \item Qwen3.5-397B-A17B-GPTQ-Int4~\citep{qwen3.5}: Apache license 2.0
\end{itemize}

\textbf{Data:}
\begin{itemize}
    \item EgoPER~\citep{lee2024error}: Custom Academic License
    \item CaptainCook4D~\citep{peddi2024captaincook4d}: Apache license 2.0
\end{itemize}

\section{Broader Impact}\label{sec:impact}
Detecting errors, in particular errors that cause harm, is very important and helps to prevent them in the future. While the zero-shot approach is a major step forward and is more practical than previous methods that require annotation of errors for training, the underlying VLM can encode societal biases present in its training data. This can lead to misclassifications of errors depending on gender, race, or other demographic attributes. The model can also hallucinate errors or provide misleading error explanations.